\documentclass[12pt]{extarticle}
\usepackage[utf8]{inputenc}
\usepackage{cite}
\usepackage{anyfontsize}
\usepackage{graphicx}
\DeclareUnicodeCharacter{3B1}{\ensuremath{\alpha}}

\title{Arbitrary Artistic Style Transfer}
\author{Weiting Li, Rahul Vyas, Ramya Sree Penta\\University of Alberta}
\date{December 17, 2022}

\begin{document}

\maketitle

\begin{center}
\fontsize{15}{0}\textbf{ Brief Abstract}
\begin{flushleft}

Arbitrary Style Transfer is a technique used to produce a new image from two images - a content image, and a style image. The newly produced image is unseen and is generated from the algorithm itself. Balancing the structure and style components has been the major challenge that other state-of-the-art algorithms have tried to solve. Despite all the efforts, it's still a major challenge to apply the artistic style that was originally created on top of the structure of the content image while maintaining consistency. In this work, we solved these problems by using a Deep Learning approach using Convolutional Neural Networks\cite{Convolu}. Our implementation will first extract foreground from the background using the pre-trained Dectron2 model from the content image, and then apply the Arbitrary Style Transfer technique that’s used in SANet\cite{SANet}. Once we have the two styled images, we will stitch the two chunks of images after the process of style transfer for the complete end piece. \end{flushleft}

\end{center}

\section{Introduction}
\medskip 

Conventional style transfer is a stroke-based rendering and image filtering method normally rely on low-level hand-created highlights. Some methods adjust the content features to match the second order statistics of the style features. Gatys \cite{Gatys} demonstrated that style patterns may be accurately captured using the statistical distribution of features extracted from pre-trained deep convolutional neural networks\cite{Convolu}. These methods define the issue as a sophisticated optimization problem, which results in significant processing cost despite the impressive outcomes.

\medskip Using a unified model, arbitrary style transfer techniques create more adaptable feed-forward architectures. AdaIN\cite{AdaIn} and DIN\cite{DIN} use conditional instance normalization and directly align the overall statistics of content features with the statistics of style features. AdaIN\cite{AdaIn} only modifies the content picture's mean and variance to match those of the style image. It successfully merges the structure of the content image with the style pattern by transferring feature statistics, but the output quality decreases as a result of  oversimplification in this method.
Through a process of whitening and coloring that uses the covariance rather than the variance, WCT\cite{WCT} converts the content features into the style feature space. The style-free decoder creates the styled image by incorporating these stylized features into a pre-trained encoder-decoder module. On the other hand, the obvious drawbacks of WCT would be that it is computationally expensive if the feature has a lot of dimensions.

\medskip Avatar-Net\cite{sheng2018avatar} uses a patch-based feature manipulation module named as style decorator, which transfers the content features into semantically nearest style features. It optimizes for minimum discrepancy between their holistic feature distributions. It uses a new hourglass network equipped with multi-scale styled adaptation that works on one feed-forward pass. Due to it's patch-based mechanism, simultaneously maintaining global and local style patterns is difficult.

\medskip Style-attentional network or SANet\cite{SANet} is closely related to the style feature decorator of Avatar-Net. For style decoration, as used in Avatar-Net\cite{sheng2018avatar}, SANet modified the self-attention mechanism to a learning soft-attention-based network. In order to preserve the content structure without losing the richness of the style, identity loss is used rather than distinct content and style losses.
However, distortion artifacts could result from the dynamic production of affine parameters in the instance normalization layer. Instead, several methods follow the encoder-decoder manner, where feature transformation and/or fusion is introduced into an autoencoder-based framework. 

\medskip GAN-based methods\cite{Domain} have been successfully used in collection style transfer, which considers style images in a collection as a domain.

\medskip In order to maintain the input's content and lessen mode collapse, contrastive learning is introduced in picture translation. CUT\cite{CUT} proposes patch-wise contrastive learning by cropping input and output images into patches and maximizing the mutual information between patches.  TUNIT\cite{CUNIT} uses contrastive learning on images with comparable semantic structures after CUT\cite{CUT}. For arbitrary style transfer tasks, the semantic similarity assumption does not hold, which causes the learned style representations to have a significant performance drop. IEST\cite{IEST} uses feature statistics (mean and standard deviation) as style priors and contrastive learning for image style transfer. Only the results that were generated were used to calculate the contrastive loss.

\medskip In this work, we propose a novel arbitrary style transfer algorithm that synthesizes high-quality stylized images in real time while preserving the content structure. This is achieved by applying the SANet to the segmented objects individually and overlay them back using the masks generated during segmentation. 

\medskip Our proposed network consists of three steps. Firstly we are segmenting the foreground and background using Dectron2 and obtain its masks as well. Secondly we apply the SANet algorithm to the segmented objects individually. Finally we overlay the segmented objects using the masks generated during segmentation. Our experiments show that our method is highly efficient where it highlights the details and outlines with an adequate amount of exposure and gave good results when there are more objects in the image.

\section{Literature Review}

One of the major challenges when doing style transfer is retention of structure while style is applied in order to keep the original intent from the content image. Alexey A. Schekalev’s paper\cite{importancemask} proposes a solution which uses an automatically generated importance mask for the content image. This mask is used to impose style with spatially varying strength, controlled by the importance mask. This paper incorporates  non-uniform stylization by using an α matrix instead of a fixed value for all spatial locations. In the paper, segmentation based importance mask generators showed the best results as per the survey results.

In 2016 Leon Gatys\cite{gats} proposed a method that uses image representations derived from Convolutional Neural Networks optimized for object recognition, which make high level image information explicit. The main idea was to optimize in the space of images to find a picture semantically reflecting content from the content image and the style of the style image. These two contradicting goals were regulated by minimizing simultaneously content loss and style loss. Coefficient α determines the strength of stylization.

In Arbitrary Style Transfer paper in Real-time with Adaptive Instance Normalization paper\cite{Normalization}, their method is a novel adaptive instance normalization (AdaIN) layer that aligns the mean and variance of the content features with those of the style features. It can transfer different regions of the content image to different styles. This is achieved by performing AdaIN separately to different regions in the content feature maps using statistics from different style inputs, but in a completely feed-forward manner. The decoder is only trained on inputs with uniform styles, but it naturally generalizes to inputs with distinct styles in various areas.

Deep feature reshuffle, which refers to spatially rearranging the position of neural activation. In Arbitrary Style Transfer with Deep Feature Reshuffle paper\cite{Reshuffle}, they provided a new understanding of neural parametric models and neural non-parametric models. Both can be integrated by the idea of deep feature reshuffle. They reshuffled the features of the style image according to the content image for style transfer. On one hand, the feature reshuffle enforces the distribution of style patterns, between the transferring result and the style image, to be globally consistent. On the other hand, a certain type of reshuffling features can help achieve locally semantic matching between images, as well as neural non-parametric models. They progressively recovered features from high-level layers to low-level ones, and trained a decoder to convert recovered features back to the image.

\medskip In Real-Time Arbitrary Style Transfer with Convolution Neural Network paper\cite{ASTCNN}, they proposed an algorithm named ASTCNN which is a real-time Arbitrary Style Transfer Convolution Neural Network. The ASTCNN consists of two independent encoders and a decoder. The encoders respectively extract style and content features from style and content and the decoder generates the style transferred image images. One style image A and one content image B is input into ASTCNN, and one image with the content of A and the style of B will be output.

\medskip In Arbitrary Style Transfer with Parallel Self-Attention paper\cite{Attention}, they introduced an improved arbitrary style transfer method based on the self-attention mechanism. A self-attention module is designed to learn what and where to emphasize in the input image. In addition, an extra Laplacian loss is applied to preserve structure details of the content while eliminating artifacts.

\medskip In Arbitrary Style Transfer with Style-Attentional Networks paper\cite{SANet}, they introduced an novel arbitrary style transfer algorithm that synthesizes high quality stylized images in real time while preserving the content structure. The SANets and decoders learn the semantic relationships between the content features and the style features by spatially rearranging the style features in accordance with the content features. At inference time, the style picture is substituted for one of the input images, and the content image is then restored to the greatest extent possible using the style features. So Identity loss helps to maintain the content structure without losing the richness of the style because it helps restore the contents based on style features.



\begin{center}
\includegraphics[scale=.8]{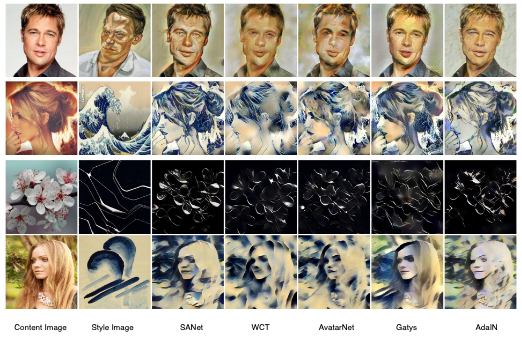}
    \scriptsize Figure 1.2 Shows visual difference in outputs when different arbitrary style transfer algorithms are compared. \cite{SANet}
\end{center}

\section{Method}

In this work, we first segmented the image into foreground and background. This Framework dynamically segments each instance in the image and produce a set of individual object images with a black background. The feedforward network, composed of SANets and decoders, learns the semantic correlations between the content features and the style features by spatially rearranging the style features according to the content features. SANet is applied to every segmented instances and finally all pieces were put back to the base image as final result. 

\begin{center}
\includegraphics[scale=.5]{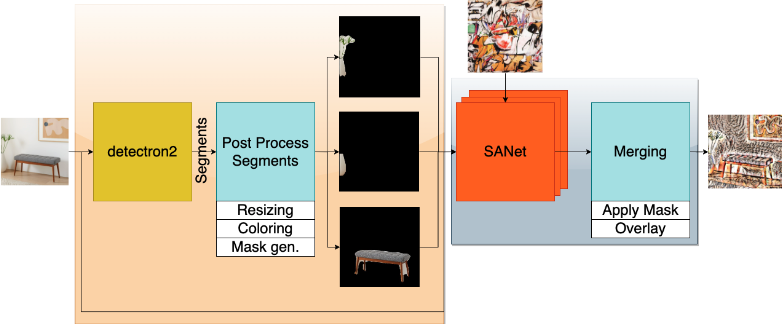}

    \scriptsize Figure 1.1 Taking a content image and a style image, we segment the object and image background; we then apply SANet to individual components, and merge the stylized output to obtain the result.
\end{center}

\medskip

\noindent\fontsize{12}{0}\textbf{3.1 Segmentation}
\medskip 
\\\indent For image segmentation, many existing techniques have proven to have outstanding performance. For the implementation of this project, our team initially plan to adopt the SoloV2\cite{SOLOv2}, a dynamic and fast instance segmentation framework that follows the SOLO (segmenting object by locations) method. However, as the project progresses, the team found a more straightforward implementation called Dectron2\cite{wu2019detectron2}, which is developed under Facebook AI research for object detection. Dectron2\cite{wu2019detectron2} is trained with COCO2017 dataset\cite{cocodataset} and utilizes Region-based Convolutional Neural Network (RCNN)\cite{renNIPS15fasterrcnn} and Feature Pyramid Network (FPN)\cite{8099589}.

\begin{center}
\includegraphics[scale=.4]{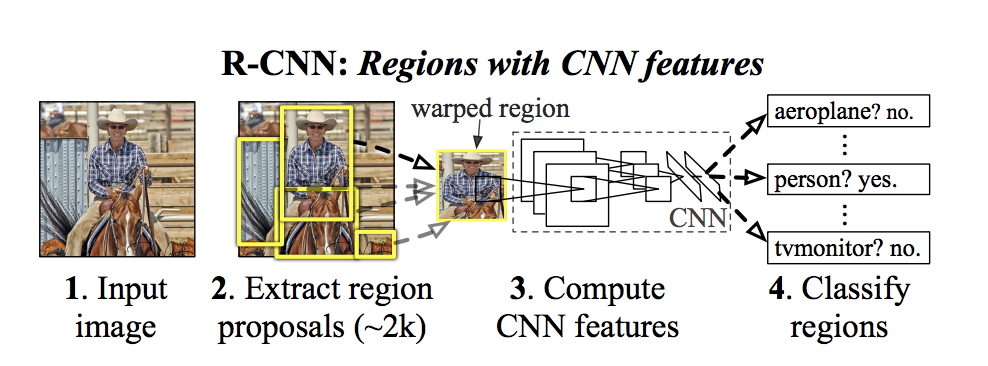}

    \scriptsize Figure 1.2 Region-based CNN start with initial sub-segmentation and used greedy algorithm to combine similar regions into large ones. Extracted features are then feed into SVM to classify objects.
\end{center}
By utilizing the masks produced by Dectron2, extracting objects from stylized and noisy images becomes much convenient with a better result.

\medskip
\indent During our implementation, we utilize some of the basic function of Dectron2, from which we obtain the segmented parts as well as their masks. For example, using an image that has several instances in it, the extracted results image can be shown below.

\begin{center}
\includegraphics[scale=.18]{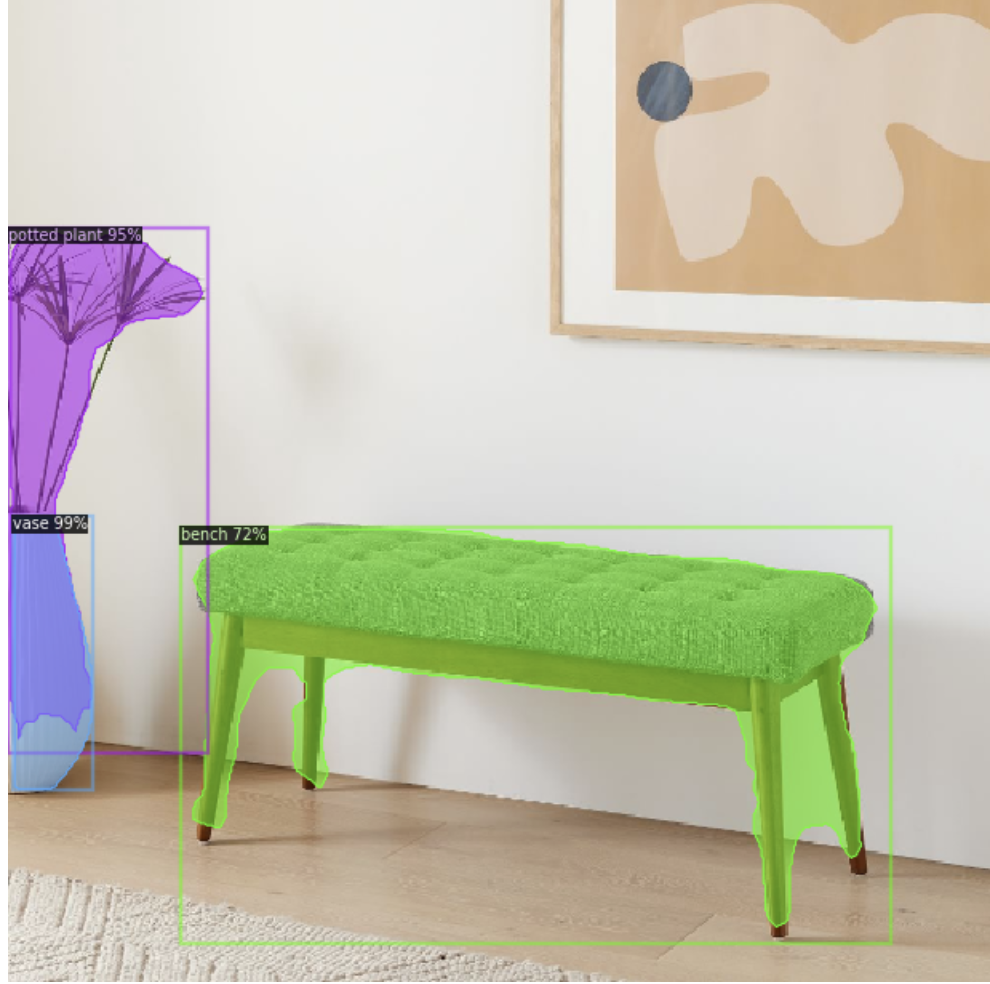}
\includegraphics[scale=.18]{graphs/segmentation/bench/Seg0_blackBG.jpg}
\includegraphics[scale=.18]{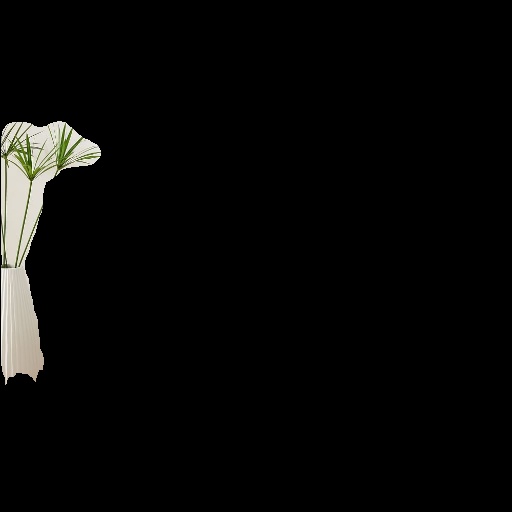}
\includegraphics[scale=.18]{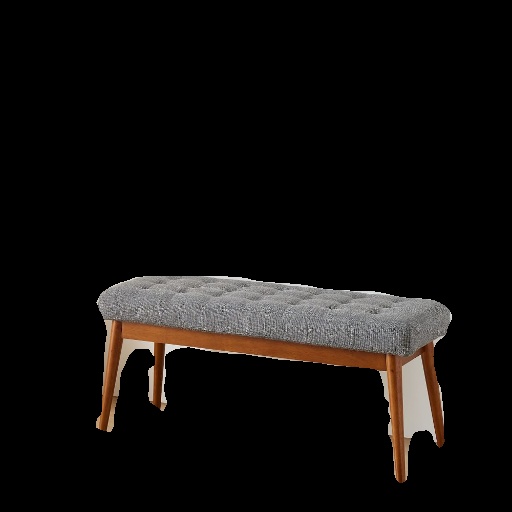}

 \scriptsize Figure 1.3 Dectron2 performs segmentation and labels individual instances accordingly. Each object can be segmented and pasted to a new image on the designated position with a black background so that the object itself can be stylized separately.
\end{center}

In the result above, we see that every segmented components were paste to images with black background. This is because we obtain the location information of the mask during segmentation stage - pasting the object to the exact location can prepare it better for the extraction after stylizing process. 

\noindent\fontsize{12}{0}\textbf{3.2 SANet}

\indent SANet is keen on stylizing image while maintaining both the global and local style pattern by using a encoder-decoder module along with a style-attentional module, as well as an unique identity loss function for leave the detailed structure untouched. During the implementation stage, it is worth noting that SANet refuses to work with alpha channel and only accepts image sizes with a power of 2, therefore we need to adjust the input image accordingly to avoid errors.

\begin{center}
\includegraphics[scale=.18]{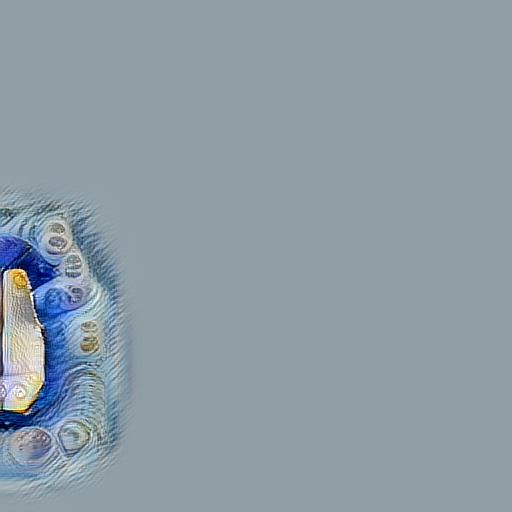}
\includegraphics[scale=.18]{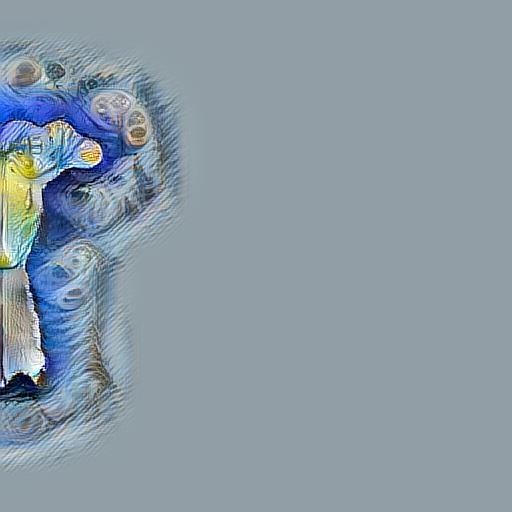}
\includegraphics[scale=.18]{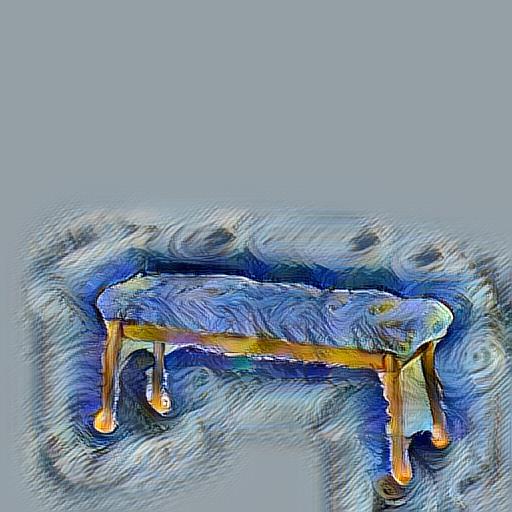}
\includegraphics[scale=.18]{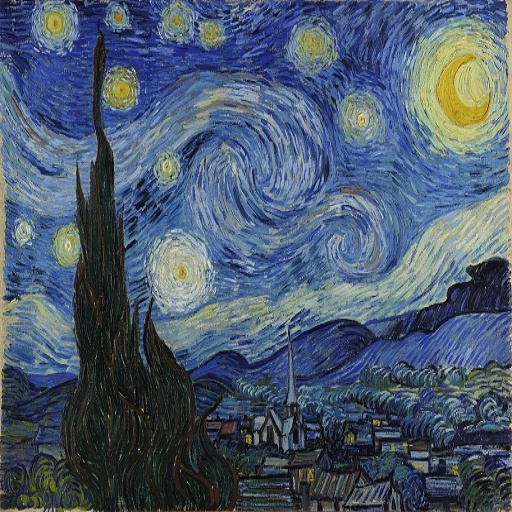}
\\ \scriptsize Figure 2.1 Applying style to the image globally makes the black backgound turned grey, which is a mere way to imitate the color cue of the original style image.
\end{center}

\indent From figure 2.1, we can see that with the result from Dectron2, applying SANet to each image yield results that highlight the desired details and outlines with an adequate amount of exposure. Under such circumstances, the image of each stylized object were treated as one single instance, therefore each of them will be stylized with all elements from the style image rather than only some of them - which is a strong contrast compared to the objects in the base image.\\
\medskip 
\noindent\fontsize{12}{0}\textbf{3.3 Combining Results}

\indent The last step of the structure is to extract each pieces and paste them on the base image. Similar to how the components were extracted initially, we use the mask of each object from the segmentation stage and apply them respectively to the stylized instances, excluding any excessive surroundings. From the figure 3.1, we can see that flower, vase and bench were paste to the base image one by one from left to right. 

\begin{center}
\includegraphics[scale=.23]{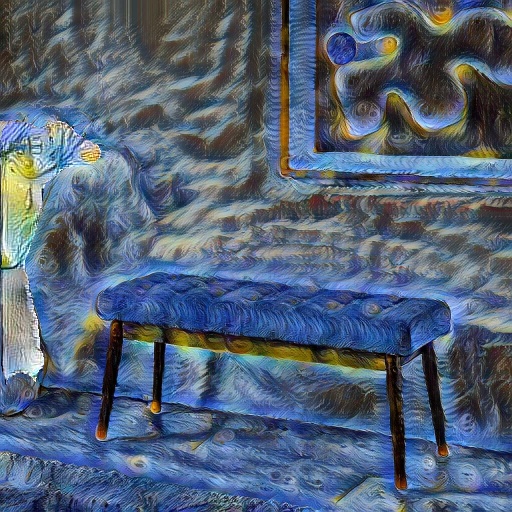}
\includegraphics[scale=.23]{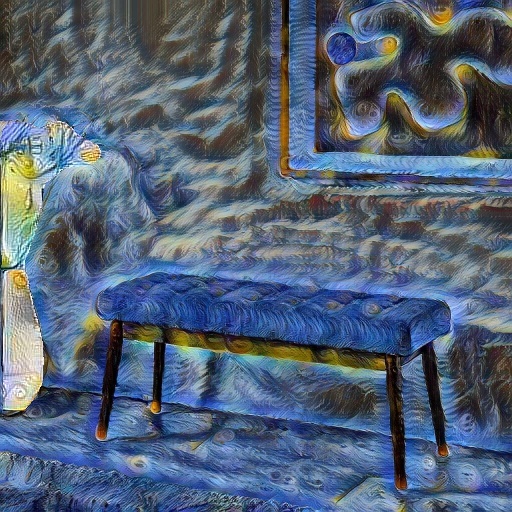}
\includegraphics[scale=.23]{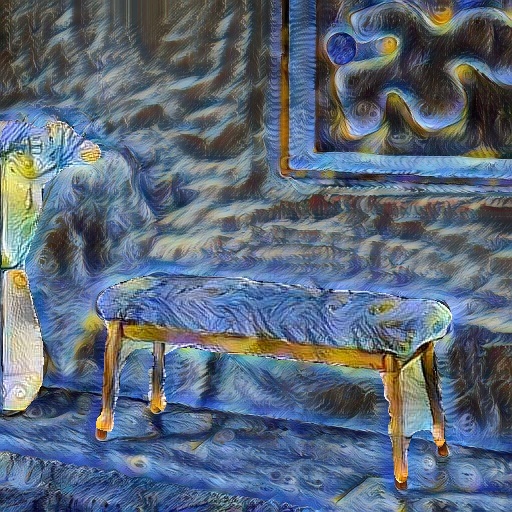}
\\ \scriptsize Figure 3.1 Here are the three steps of pasting individual instances, which has been stylized separately, back to the base image. First the flower, then the vase and finally the bench.
\end{center}

\section{Result}
\begin{center}
\includegraphics[scale=.09]{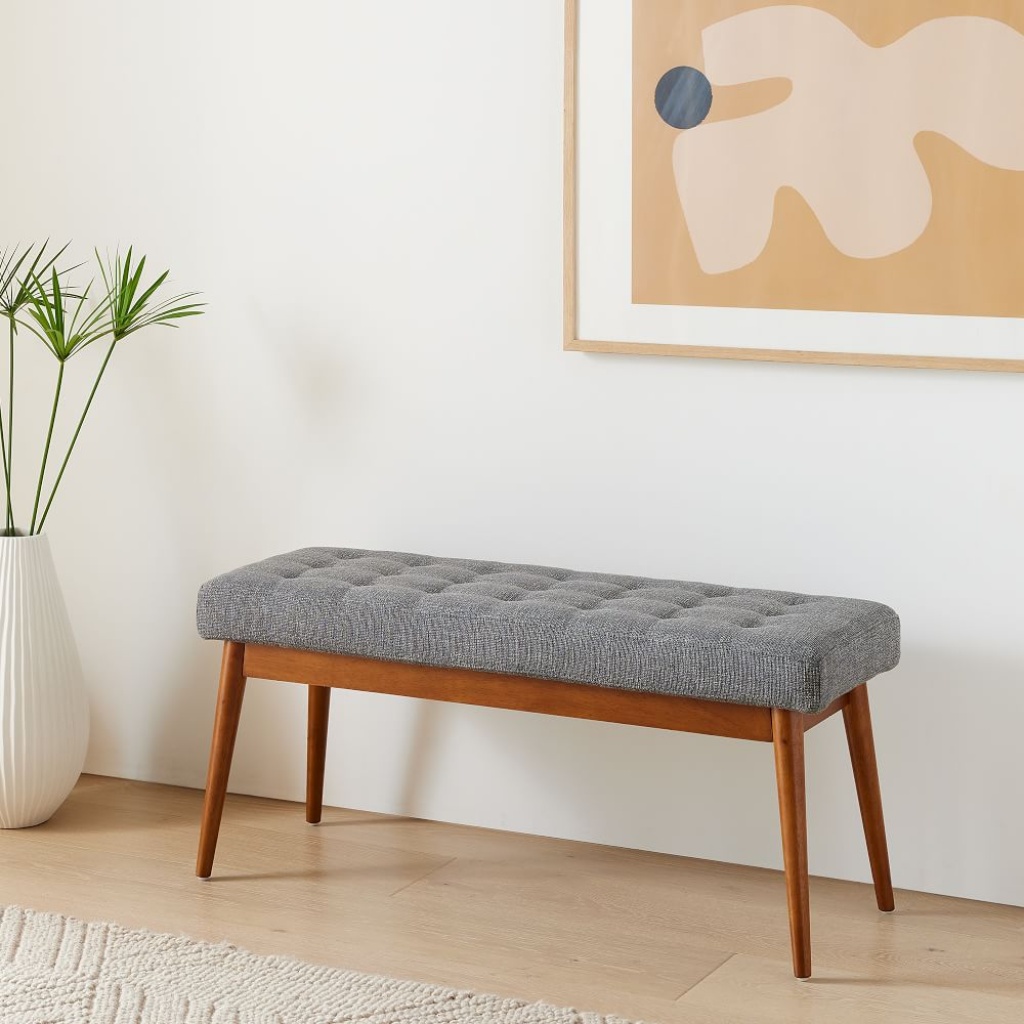}
\includegraphics[scale=.18]{graphs/segmentation/Starry.jpg}
\includegraphics[scale=.09]{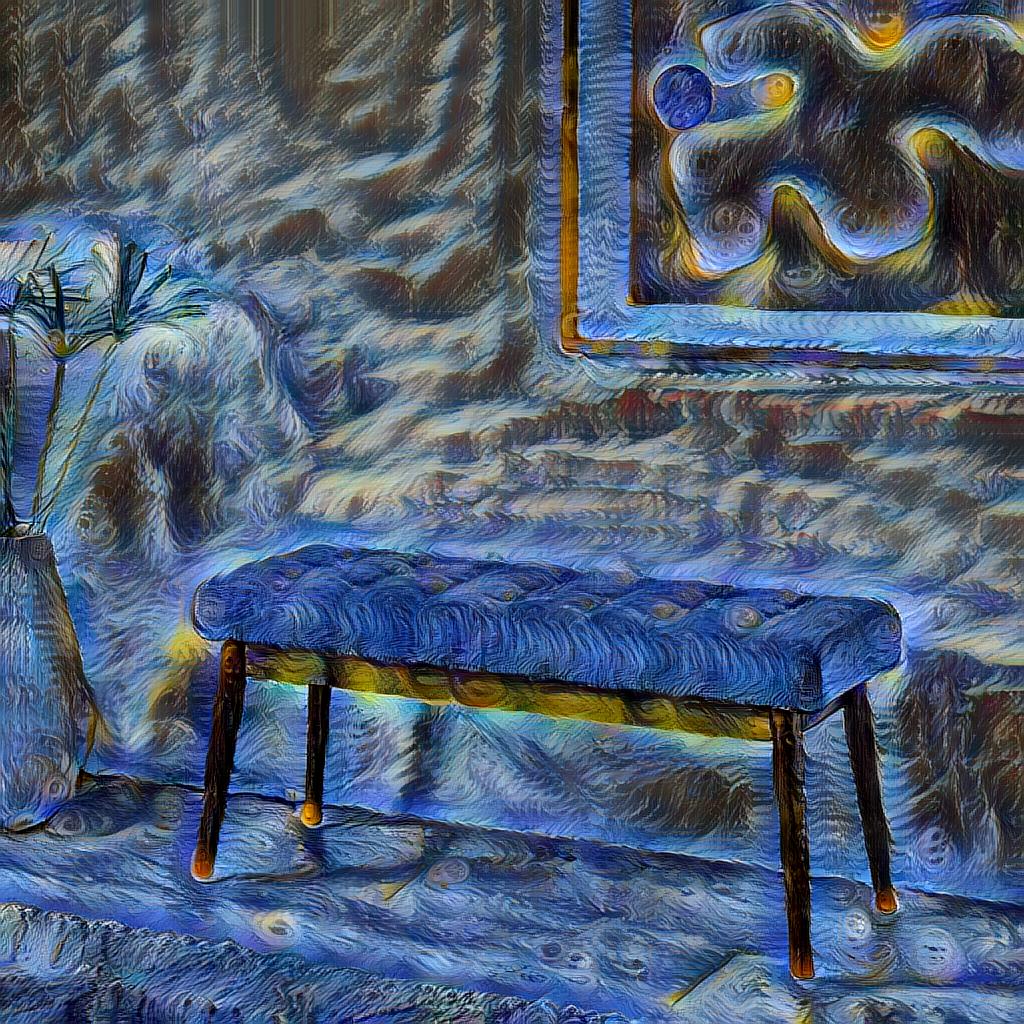}
\includegraphics[scale=.18]{graphs/segmentation/bench/Result_2.jpg}

\includegraphics[scale=.18]{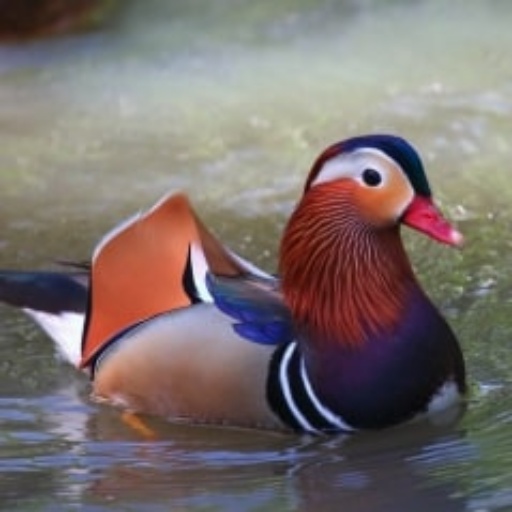}
\includegraphics[scale=.18]{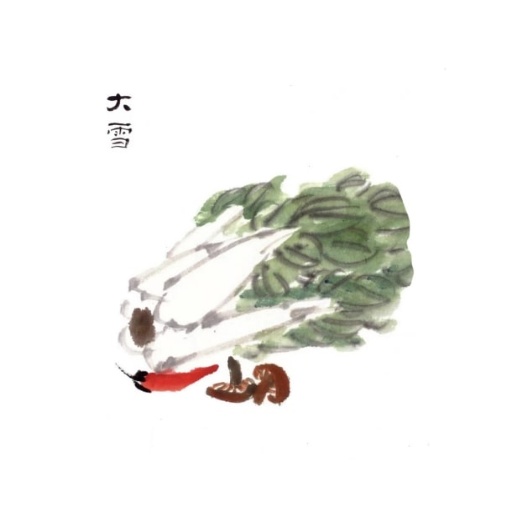}
\includegraphics[scale=.18]{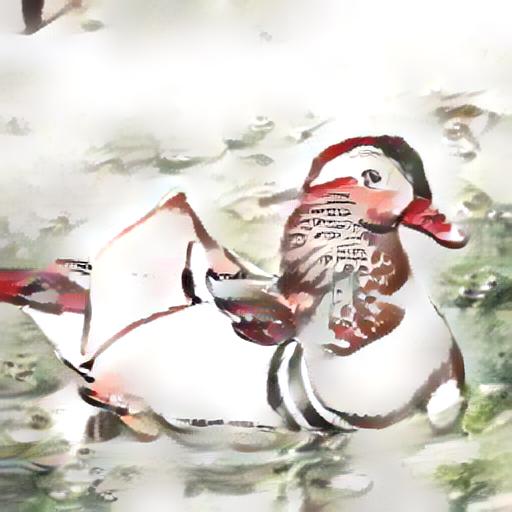}
\includegraphics[scale=.18]{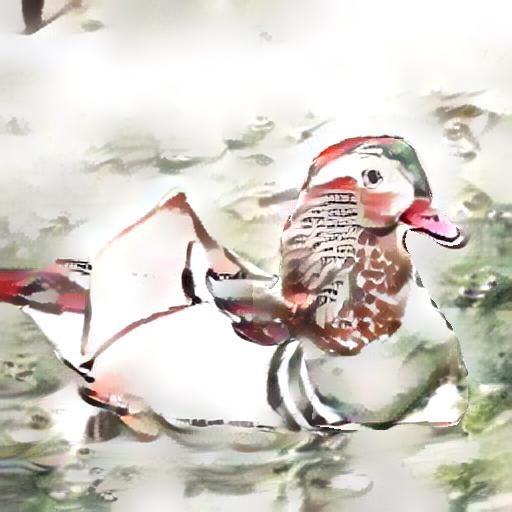}

\medskip 

\includegraphics[scale=.18]{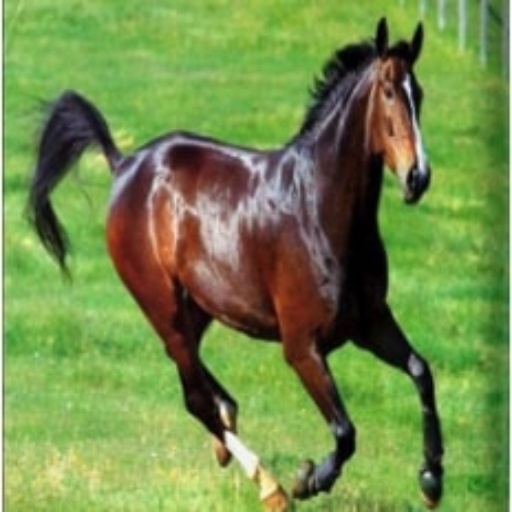}
\includegraphics[scale=.18]{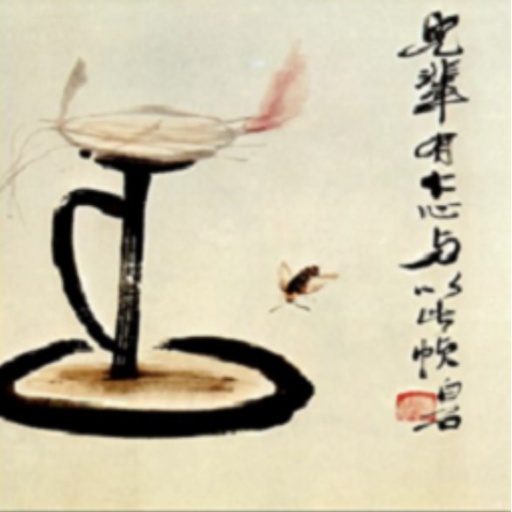}
\includegraphics[scale=.18]{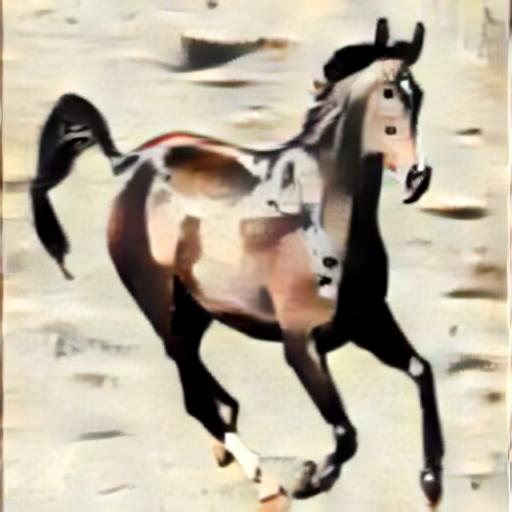}
\includegraphics[scale=.18]{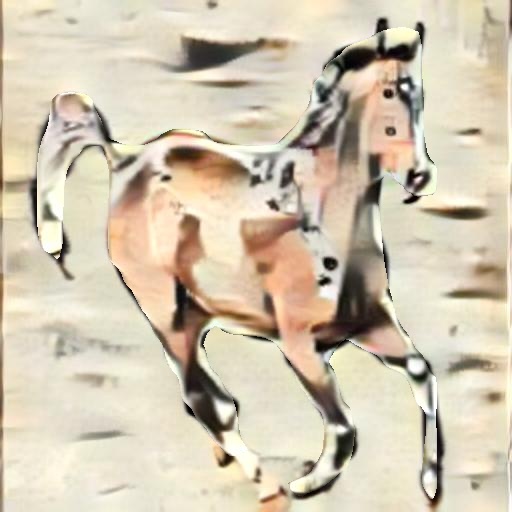}

\scriptsize Figure 4.1 Images on the first column are original; the second column is the style images, the third are the SANet results, and the last are results from our structure.
\end{center}

\medskip 
\indent By combining the results from segmentation and style transfer, the quality of the end piece is more or less a question of personal preference. However, from figure 4.1, we can tell that for the bench image, our result appears to have a more vibrant color to different components, thus making each object more distinct comparing to their appearances in the SANet result; the duck image are almost the same while our result shows more texture and focus on the details, making the instance blends well yet stay separated from the background. For the horse image, while some may argue the color shifts destroys the texture of the horse, it presents another dynamics of color scheme in style transfer and our result continues to show a remarkable performance in this case. 

\indent The main idea is that our structure works comparatively well in multi-object scene. For single-object scene, backgrounds are sedentary and can be considered as an image without any noise, therefore it was easier to perform style transfer on such image as some factors like noise can be excluded. However, in the case where multiple objects exist, applying segmentation before the style transfer can help the image to produce a clean boundary for each object, which essentially yields better results for multi-object scenes.

\section{Discussion}

The novel approach of applying segmentation first and then applying SANet on each segment outperforms SANet for specific content images, especially those including multiple objects so the content form is not lost. However, there are a few improvements. Talking to our professor \& mentors, we received feedback about comparing multiple segmentation algorithms. The results from detectron2 did not classify all the objects in the scene. In our Figure 1.3, we also have a photo frame which was undetected. Another qualitative result we can see is that the images have less contrast. The reason for this could be that the style is being applied locally. One solution suggested for this is to apply similar segmentation to style image and having a mapping created between the style and content image which will then be applied proportionally. Another improvement could be made on the edges. The edges must be smooth and for that we could apply different image processing techniques to handle edges better. Understanding these, we want to also explore the idea of creating content aware (importance) masks so that style coefficient varies based on the content importance.

\bibliographystyle{plain}
\bibliography{main}

\end{document}